\documentclass[final]{cvpr}

\usepackage{times}
\usepackage{epsfig}
\usepackage{graphicx}
\usepackage{amsmath}
\usepackage{amssymb}

\usepackage{soul}
\usepackage{bm}
\usepackage{algorithm}
\usepackage{algorithmicx}
\usepackage[noend]{algpseudocode}
\usepackage{mathtools}
\usepackage{caption}
\usepackage{subcaption}
\usepackage{xcolor}
\usepackage{enumitem}
\usepackage{booktabs}
\usepackage{xspace}

\usepackage[pagebackref=true,breaklinks=true,colorlinks,bookmarks=false]{hyperref}

\def\cvprPaperID{****} 
\def\confYear{CVPR 2021}

\definecolor{darkgreen}{rgb}{0,0.45,0}
\definecolor{purple}{rgb}{0.5,0,0.5}

\newcommand{\giscard}[1]{}
\newcommand{\dian}[1]{}
\newcommand{\suzie}[1]{}
\newcommand\minisection[1]{\vspace{1mm}\noindent \textbf{#1}}

\newcommand{\method}{$\text{ReAL}$\xspace}
\newcommand{\ours}{$\text{ReAL}$\xspace}

\newcommand{\object}{$\text{Object-level}$\xspace}
\newcommand{\image}{$\text{Image-level}$\xspace}

\newcommand{\xview}{$\text{xView}$\xspace}
\newcommand{\coco}{$\text{COCO 2017}$\xspace}

\algnewcommand\algorithmicforeach{\textbf{for each}}
\algdef{S}[FOR]{ForEach}[1]{\algorithmicforeach\ #1\ \algorithmicdo}

\begin{document}

\title{Region-level Active Detector Learning}

\author{Michael Laielli, Giscard Biamby, Dian Chen, Ritwik Gupta, \\Adam Loeffler, Phat Dat Nguyen, Ross Luo,
Trevor Darrell, Sayna Ebrahimi\\
UC Berkeley\\
{\tt\small \{laielli,gbiamby,dian,ritwikgupta,aross.1311,phatdatn,ross.luo,trevordarrell,sayna\}@berkeley.edu}
}

\maketitle

\begin{abstract}
    Active learning for object detection is conventionally achieved by applying techniques developed for classification in a way that aggregates individual detections into image-level selection criteria. This is typically coupled with the costly assumption that every image selected for labelling must be exhaustively annotated. This yields incremental improvements on well-curated vision datasets and  struggles in the presence of data imbalance and visual clutter that occurs in real-world imagery. Alternatives to the image-level approach are surprisingly under-explored in the literature. In this work, we introduce a new strategy that subsumes previous Image-level and Object-level approaches into a generalized, Region-level approach that promotes spatial-diversity by avoiding nearby redundant queries from the same image and minimizes context-switching for the labeler. We show that this approach significantly decreases labeling effort and improves rare object search on realistic data with inherent class-imbalance and cluttered scenes.
\end{abstract}
\section{Introduction}

Training a visual object detector often requires massive amounts of  bounding box labels in order to achieve desired performance levels. This data is expensive to collect since the labeling process requires multiple stages of exhaustive searches for the presence of multiple categories and instances. Active learning is an attractive mitigation strategy that aims to label only the most informative samples, achieving a desired performance with less labels. However, applying active learning to visual object detection is not straightforward.

Existing active learning strategies for object detection fall into two distinct approaches.   The first, and most prevalent in the literature, takes an \textit{Image-level} approach that sends a query to the ``oracle'' labeler in the form of a whole-image for exhaustive labeling, typically selected based on object prediction scores aggregated within each image ~\cite{roy2018deep, brust2018active, aghdam2019active}. This approach becomes very inefficient when valuable rare objects occur in images that also contain hundreds or thousands of common objects.

\begin{figure}[!t]
\includegraphics[width=0.48\textwidth]{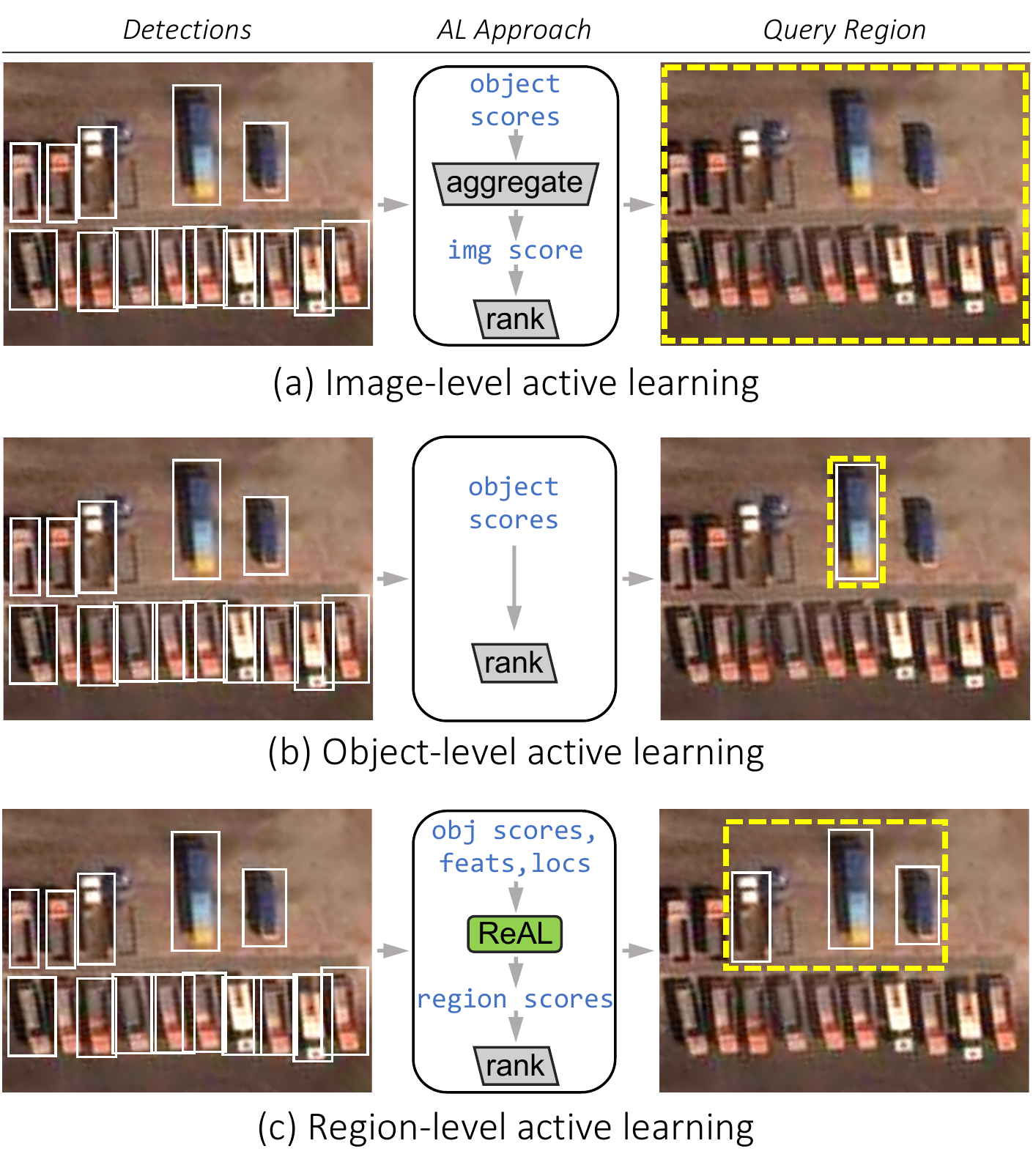}

\caption{ A satellite image example of parked vehicles. a) Image-level active detector learning aggregates the informativeness (e.g. entropy) of detected objects into a single score before ranking amongst other unlabeled images and selecting the entire image area as a query to the oracle. b) Object-level active detector learning ranks, selects, and queries on single objects only. c) Region-level active detector learning (ours) additionally utilizes object features and locations to define and rank regions that are informative in terms of the uncertainty and diversity of the  objects they contain.
}
\label{fig:real}
\end{figure}

While \image minimizes the total number of images scanned, it is vulnerable to intra-image redundancy: objects within the same image are likely to be similar and therefore redundant in terms of little-to-no added information. The \textit{Object-level} approach, foregoes exhaustive labeling and sends a single candidate object bounding box as a query to the oracle~\cite{wang2018towards, desai2020towards}. \object increases diversity by breaking the exhaustive labeling assumption, but incurs several costs such as introducing many false negatives to the labeled pool of images, increasing visual search, and maximizing context switching required of the labeler.

In this work, we propose \method, a \textit{Region-level} Active detector Learning approach that selects regions of high informativeness that contain diverse groups of objects while minimizing the query area. \method, depicted in Figure \ref{fig:real}, finds where neighboring candidate objects have low similarity to one another in terms of feature embeddings from the detection model.

\method is agnostic to both the scoring function and the selection strategy, and is thus easily applied to typical active learning methods comprised of a scoring function, selection strategy, or both.
Our approach avoids label redundancy without drastically increasing false negatives or total pixels scanned by not being constrained to either extreme of image- or object-level fixed approaches.

In extreme cases of images densely packed with large objects or sparsely, with small objects, \method can equate to the Image- or Object-level approaches, respectively.

We evaluate \method against the Image- and Object-level baselines with a suite of previously reported  Active Learning for detection methods.  We additionally present a novel  extension of a recent Minimal Active Learning \cite{MAL} method to handle the detection setting, which we call D-MAL, and which incorporates both uncertainty and diversity as measures of informativeness for detection active learning. 

Our natural environment is full of clutter: diverse sets of object classes comprise everyday scenes. However, curated vision datasets aim to reduce clutter in each instance. To demonstrate the efficacy of our method, we evaluate across datasets representing both ends of this spectrum: The classic object detection benchmark \coco with low object clutter, and the xView dataset containing highly varying levels of object clutter, class imbalance, and label noise.
We confirm that \ours and \object performs similarly on \coco, as expected due to the homogeneity introduced by heavy dataset curation.
On xView, our \ours approach dynamically finds image regions with diverse, high-entropy objects to reduce the labeling effort by up to $60\%$ as compared to \object and \image approaches with an entropy-based method.  We also show that \ours combined with our novel D-MAL method outperforms across multiple methods in terms of detection accuracy and object search with relative improvements of up to 24$\%$ and 41$\%$, respectively.

\section{Related Work}

Although active learning for object detection is a long-standing problem~\cite{abramson2004active,kapoor2007active,bietti2012active,chen2014active}, it has remained relatively under-explored compared to active learning for image classification \cite{vaal,MAL, beluch2018ensemble, coreset, wang2014label,suggestiveannotation, preclustering,ash2020deep,gorriz2017costeffectivemelanoma,wang2017cost}. Sampling strategies developed in the literature either request for full-image annotations~\cite{roy2018deep, brust2018active, aghdam2019active} or select individual objects to be labeled~\cite{wang2018towards, desai2020towards}.

\begin{figure*}[t]
\centering\vspace{1em}
\includegraphics[width=0.95\textwidth]{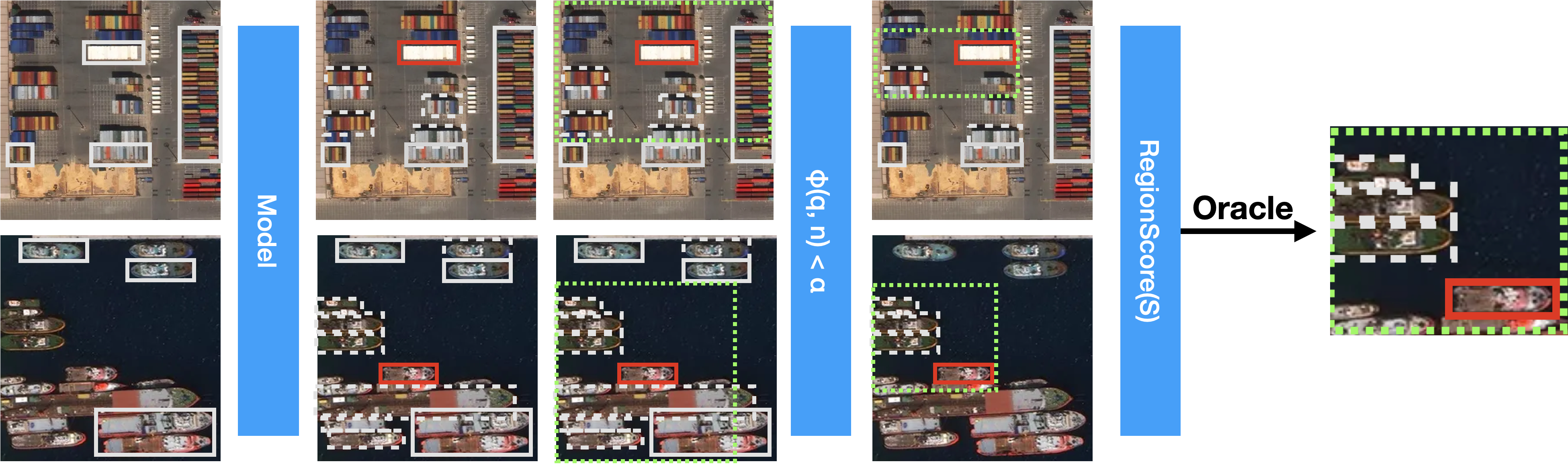}

\caption{
Conceptual illustration of our proposed method, Region-level Active detector Learning (\method). A set of partially labeled images ($X_P$) are passed through an object detection model which produces candidate object bounding boxes. A query object is selected around which a context window is dynamically constructed (Eq. \ref{eq:context_window}). Other candidate objects within the context window are filtered by similarity to the query. Regions are then scored by Eq. \ref{eq:region_score}, the highest scoring of which are sent to the oracle for labeling.}
\label{fig:real_blackbox}
\end{figure*}

Image-level approaches often follow algorithms that were originally proposed for classification settings where they aggregate results computed for different instances in an image to decide whether or not an image should be labeled. Roy et al. \cite{roy2018deep} used a query by committee approach~\cite{qbc} to query the image with the largest disagreement between the convolutional layers in the object detector backbone. Brust et al. \cite{brust2018active} used uncertainty measures on detected bounding boxes to compute a per-image aggregated score.  Aghdam et al. \cite{aghdam2019active}
 proposed an Image-level score that accumulated pixel-level importance scores such that an image with the highest number of false-positive and false negative
predictions would be sent to the oracle. Desai et al. \cite{desai2019adaptive} proposed an adaptive learning strategy where at each cycle, they first used weakly-labeled data received from the labeler to optimize the model and then switched to a stronger form of supervision as required when further training the model. Kao et al. \cite{kao2018localization} proposed two metrics to measure object informativeness including `localization tightness' and `localization stability' where the former is based on the overlapping ratio between the region proposal and the final prediction and the latter is based on the variation of predicted object locations when input images are corrupted by noise.

Instance-level methods were introduced to mitigate the costs of full-frame labeling in Image-level strategies. Wang et al. \cite{wang2018towards} proposed a Self-supervised Sample Mining using pseudo-labeling for enhancing  the object detector via the introduced cross image validation, i.e., pasting these proposals into different labeled images to comprehensively measure their values under different image contexts. Desai et al. \cite{desai2020towards} used two metrics for querying bounding boxes to be labeled including uncertainty associated with classification predictions and coreset method~\cite{coreset} that computed distance based on bounding box feature representations.

\section{Region-level Active Learning (ReAL)}\label{sec:method}

We consider the problem of learning a detection model by efficiently selecting the most informative 
image areas to be labeled by an oracle. We start with an initial pool of images 
denoted as  $\mathcal{L}=\{(x^i_L,y^i_L)\}_{i=1}^{N_L}$,
where $y^i_L=\{O_j\}_{j=1}^{M_i}$ is the set of annotations of each image. For a fully-annotated dataset, the initial pool is constructed by randomly sampling from all available annotations, and taking all the images associated with the annotations. We uniformly sample $p\%$ of all annotations per category with a hard cap $k$ on the number of instances. This reflects our prior knowledge of a dataset in that we have more access to the common categories and less to the rarer categories, while avoiding exhausting any category in the beginning. The remaining images constitute the pool of unlabeled data, which is denoted as $X_U=\{x^i_U\}_{i=1}^{N_U}$ where our goal is to populate a fixed sampling budget, $b$, using an acquisition strategy to query instances from the unlabeled pool ($x_U \sim X_U$). 

Given an initial labeled pool of images and any informativeness scoring function, $\psi$, that predicts the potential learning value of a candidate object, ReAL can proceed.  \method takes detection model predictions as input. We consider these predictions to be \textit{candidate objects} for potential verification by the oracle.

\method introduces optimal query regions. The main idea is to send to the oracle only the image regions that contain the most diverse candidate objects with the highest potential learning value. ReAL makes two main contributions compared to Image-level and Object-level baseline approaches, algorithms 1 and 2.  The first component, algorithm 3 lines 7 and 8, defines regions with low candidate object similarity. The second component, line 2, scores regions so that those with high informativeness and low candidate object similarity score higher.  These components are described in detail in the following sections.


\begin{figure*}[!t]
	\centering
\begin{minipage}[t]{0.47\textwidth}
\begin{algorithm}[H]
\begin{algorithmic}[1]
    \Require unlabeled images $X_U$, fully labeled images $(X_L,Y_L)$, labeling budget $b$, detection model $\mathcal{D}$, informativeness function $\psi(\cdot)$
    
    \State $\mathcal{D} \gets \text{train}(\{(X_L,Y_L)\}) $
    \Repeat
        \State $x^* \gets  \arg \max_{x \in X_U} \frac{1}{|\mathcal{D}(x)|}\sum_{q \in \mathcal{D}(x)}(\psi(q))$
        \State $Y_O \leftarrow Y_O \cup \text{Oracle}(x^*)$
    \Until{$|Y_O| = b$}
    \State \Return {$Y_O$}
\end{algorithmic}
\caption{\textsc{Image-level Baseline Approach}}\label{algo:image-level}
\end{algorithm}

\vspace*{-\baselineskip}

\begin{algorithm}[H]
\begin{algorithmic}[1]
    \Require unlabeled images $X_U$, partially labeled images $(X_P,Y_P)$, fully labeled images $(X_L,Y_L)$, labeling budget $b$, detection model $\mathcal{D}$, informativeness function $\psi(\cdot)$
    
    \State $\mathcal{D} \gets \text{train}(\{(X_P,Y_P) \cup (X_L,Y_L)\}) $
    \Repeat
        \State $q^* \gets  \arg \max_{q \in \mathcal{D}({\{ X_U \cup X_P\}})} \psi(q)$
        \State $Y_O \leftarrow Y_O \cup \text{Oracle}(q^*, x_{q^*})$
    \Until{$|Y_O| = b$}
    \State \Return {$Y_O$}
\end{algorithmic}
\caption{\textsc{Object-level Baseline Approach}}
\label{algo:object-level}
\end{algorithm}

\end{minipage}
\hfill
\begin{minipage}[t]{0.50\textwidth}
\begin{algorithm}[H]
\begin{algorithmic}[1]
    \Require unlabeled images $X_U$, partially labeled images $(X_P,Y_P)$, fully labeled images $(X_L,Y_L)$, labeling budget $b$, detection model $\mathcal{D}$, cosine similarity function $\phi(\cdot,\cdot)$, informativeness function $\psi(\cdot)$

    \Function{RegionScore}{$R$}
        \State 
        \Return $\psi(q_R) + \sum_{n \in R \setminus \{q\}}(\psi(n)(1 - \phi(q_R,n)))$
    \EndFunction
    \Function{ObjectsInContextWindow}{$O$,$q$}
        \State $q_{aw},q_{ah}  = \beta (1 - \frac{q_{bw}}{q_{iw}})^{\beta}q_{bw}, \beta (1 - \frac{q_{bh}}{q_{ih}})^{\beta}q_{bh}$
        \State  $q_{ax}  = \{x \mid q_{bx} - \frac{q_{aw}}{2} < x < q_{bx} + \frac{q_{aw}}{2}\}$
        \State  $q_{ay}  = \{y \mid q_{by} - \frac{q_{ah}}{2} < y < q_{by} + \frac{q_{ah}}{2}\}$

        \State \Return $\{o \in O\setminus q \mid o_{bx} \subseteq q_{ax} \land o_{by} \subseteq q_{ay} \}$
    \EndFunction
    
    \State $\mathcal{D} \gets \text{train}(\{(X_P,Y_P) \cup (X_L,Y_L)\}) $
    \State $O \gets \mathcal{D}({\{ X_U \cup X_P\}})$
    \State $S \gets \emptyset$
    \ForEach {$q \in O $}
        \State $N \gets \text{ObjectsInContextWindow}(O,q)$
        \State $R \gets \{q\} \cup  \{n \in N \mid \phi(q,n) < \alpha\}$
        \State $S \gets S \cup \{R\}$
    \EndFor
    \Repeat
        \State $R^* \gets  \arg \max_{R \in S} \text{RegionScore}(R)$
        \State $Y_O \leftarrow Y_O \cup \text{Oracle}(R^*, x_{R^*})$
    \Until{$|Y_O| = b$}
    \State \Return {$Y_O$}
\end{algorithmic}
\caption{\textsc{ReAL Approach}}
\label{algo:real}
\end{algorithm}

\end{minipage}

\vspace{5pt}
\scriptsize{\textit{Algorithm 1, 2 and 3 definitions}}
\vspace{1pt}
\hrule
\vspace{1pt}
\scriptsize{$q$$:=$$\text{the \textit{query} candidate object}$ $\mid$ $R$$:=$$\text{Region; set of candidate objects}$ $\mid$ $S$$:=$ $\text{Set of regions}$ $\mid$ $n_{bx}$$:=$$\text{Set of pixel x-coords for bounding box } n_b$ $\mid$ $q_{ax}$$:=$$\text{set of pixel x-coords for local image area } q_a$ $\mid$ $\mathcal{D}(\bullet)$$:=$ $\text{model output for input } \bullet$ $\mid$$N$$:=$ $\text{Set of candidate objects neighboring } q$ $\mid$ $x^*$$:=$ $\text{image selected for labeling}$ $\mid$ $Y_O$$:=$ $\text{Labels returned by oracle}$ $\mid$ $\text{Oracle}(\bullet)$$:=$ $\text{True labels from oracle for } \bullet$ $\mid$ $x_{R^*}$$:=$ $\text{image containing } R^*$ }

\vspace{2pt}

\hrule

\end{figure*}

\subsection{Region finding}

To find regions dense with diverse objects, we start with each candidate object, $q$, produced by $\mathcal{D}(\cdot)$ on all images that have not been fully labeled, $\{ X_U \cup X_P \}$.  A single candidate object consists of category probabilities $q_c$, bounding box coordinates $q_b$, and a penultimate-layer feature vector $q_f$, all of which correspond to a single object prediction made by the detection model. A region $R$, is defined by a set of candidate objects. Initially, each $q$ is the only candidate object in its corresponding region, $R \gets \{q\}$. More candidate objects are added to $R$ by considering the cosine similarity between $q_f$ and its neighboring candidate objects ${\phi(q,n)}_{n\in N}$ where $\phi(q,n) = \frac{q_f \cdot n_f}{\lVert q_f \rVert \lVert n_f \rVert}$.  A neighboring candidate object is added to $R$ if its similarity score is less than a threshold, $\alpha$, which we set to 0.5 in our experiments without tuning. 
Given an image $x$, a detector $\mathcal{D}$, and a query candidate object $q$, a region $R$ is defined as

\vspace{-5pt}

\begin{equation}\begin{gathered}
R \gets  \{q\} \cup  \{n \in N \mid \phi(q,n) < \alpha \}
\end{gathered}\end{equation}

The neighboring candidate objects $N$ are determined by a context window surrounding the object.  We define this rectangular area $q_a$ to be centered and dependent on the size of $q_b$ as well as the size of the respective image $q_i$. The width and height of $q_a$ are

\vspace{-5pt}

\begin{equation}
\label{eq:context_window}
\begin{gathered}
q_{a_w},q_{a_h}  = \beta (1 - \frac{q_{b_w}}{q_{i_w}})^{\beta} q_{b_w},\beta (1 - \frac{q_{b_h}}{q_{i_h}})^{\beta} q_{b_h}
\end{gathered}\end{equation}

The $\beta$ parameter represents the number of times by which the boundaries of the smallest box size should be extended.  For example, a small box should yield a box-image ratio close to zero, thus $\alpha_q$ achieves the nearly full value of $\beta$, while a larger box attains a smaller fraction.

\subsection{Region Scoring}

Once a region of dissimilar predictions has been established, the region boundaries are determined by drawing the minimum bounding box that contains all $q_b \in R$. Then the region is scored according to an accumulation of the informativeness and similarity of each query-neighbor pairing.
\begin{equation}
\label{eq:region_score}
    \text{RegionScore}(R) = \psi(q_R) + \sum_{n \in R \setminus \{q\}}(\psi(n)(1 - \phi(q_R,n)))
\end{equation}

The term right of the addition subtracts the cosine similarity between query $q_R$, and neighbor $n$, from 1 to reduce the score in the case of high similarity. This value is weighted by the informativeness of the neighbor $\psi(n)$.  Then it's summed across all of the candidate objects in the region excluding the query candidate object, $n \in R \setminus \{q\}$.  Finally, the informativeness of the query candidate object is added to attain the region score.

\begin{figure*}[t]
	\centering
	\includegraphics[width=\linewidth]{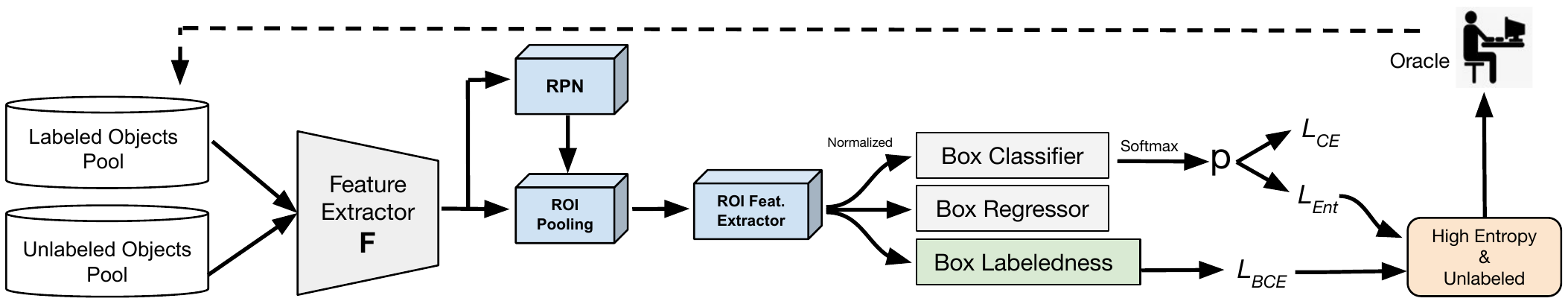}
	\vspace{-10pt}
	\caption{Our D-MAL pipeline using a Faster-RCNN detector. Both labeled and unlabeled data feed into the Faster-RCNN's backbone architecture $F$ as the feature extractor. We have added an extra head to the original Faster-RCNN architecture which is trained to predict \textit{labeledness} which refers to whether an object has been labeled or not. 
	The original classifier head is also replaced with a normalization layer followed by a cosine similarity-based classifier $C$ which is trained to maximize entropy on unlabeled objects, whereas $F$ is trained to minimize it. We have evaluated D-MAL with Image-level, Object-level, and our proposed Region-level selection strategies.}
	\label{fig:mal-detection}
	\vspace{-10pt}
\end{figure*}

\section{Minimax Active Learning for Detection} \label{sec:mal4det}
In this section, we describe our novel extension of the Minimax Active Learning method (MAL) \cite{MAL} for object detection, D-MAL, depicted in Figure \ref{fig:mal-detection}.  MAL is a semi-supervised active learning method which is originally introduced for classification and segmentation. It has the ability to query samples based on uncertainty using entropy and/or diversity using a classifier that is designed to discriminate the most \textit{different} samples from those already seen in the feature space. We use this method to analyze the uncertainty and diversity role in active learning for object detection in conjunction with our /method approach (see Section \ref{sec:approaches}). 

Figure \ref{fig:mal-detection} illustrates D-MAL architecture for detection which has three major building blocks: \textbf{(i)} a feature encoder ($F$) that attempts to minimize the entropy of the unlabeled data for better clustering. \textbf{(ii)} a distance-based classifier head that adversarially learns per-object weight vectors as prototypes and attempts to maximize the entropy of the unlabeled data \textbf{(iii)} a \textit{labeledness} head that takes the extracted ROI features as input and predicts which pool each object belongs to. D-MAL leverages both the labeled and unlabeled objects by performing the minimax entropy optimization in a semi-supervised learning fashion which is easy to train and does not require labels. 

The intuition behind D-MAL is that while the feature extractor minimizes the entropy by pushing the unlabeled objects to be clustered near the prototypes, the classifier increases the entropy of the unlabeled objects, and hence makes them similar to all class prototypes. On the other hand, they both learn to classify the labeled objects using a cross-entropy loss, enhancing the feature space with more discriminative features. We train the \textit{labeledness} head along with the classifier and the box regressor heads to predict whether an object has been labeled or not for each box. In an Object-level approach, D-MAL queries objects which (i) are predicted as unlabeled with high confidence by the \textit{labeledness} head and (ii) achieved high entropy by the classifier will be sent to the oracle. In \method approach however, the labeledness and entropy scoring functions are used as a basis to find and score multi-object regions.  In terms of Algorithm 3, the D-MAL informativeness function is a combination of the labelness and entropy components: $\psi(\cdot) = \psi_l(\cdot) + \psi_\mathcal{H}(\cdot)$. These informativeness scores are combined with measures of similarity between object candidates that are computed as the cosine similarity between normalized feature embeddings extracted from the Faster-RCNN model by the ROI Feature Extractor depicted in Figure \ref{fig:mal-detection}.


\section{Experimental setup}
In this section we review the details of our experimental setting, benchmark datasets, and baselines used in our evaluation as well as the implementation details.

\subsection{Datasets} \label{sec:dataset}
We have evaluated our \method approach on \textbf{\xview} \cite{xview} as one of the largest publicly available datasets of overhead imagery, as well as \textbf{\coco}, \cite{lin2014microsoft} which is a common object detection benchmark. Below we describe the qualities of xView that make it a good benchmarking set for Active Learning for cluttered scenes and contrast it with \coco.  We also provide details on pre-processing, choice of performance metrics, and training hyper-parameters for both datasets. License and privacy details for \xview and \coco are discussed in \cite{xview,lin2014microsoft}.

\minisection{\xview~\cite{xview}.} \xview consists of $846$ large satellite images that range from $2500$ to $3500$ pixels$^2$. $60$ classes of objects are exhaustively labeled with bounding boxes for a total of $\sim1$M labeled object instances.  The class-level breakdown of these instances follows a long-tail distribution, with two categories numbering in the hundreds of thousands, and many others in the hundreds or even tens. By comparison, all categories of the \coco dataset have at least 1000 instances.

\minisection{Data pre-processing for \xview.} In our experiment, we make some adjustments to the \xview dataset for the purposes of data processing and evaluation.  First, we join some child classes so that low-shot categories have enough instances for robust evaluation, which yields 35 total object categories. Second, we break the large images into smaller $512 \times 512$ tiles so that the small batches will fit in GPU memory during training. Our version has $19,631$ image tiles of size $512 \times 512$ and $418,217$ total labels in the train set and $6,910$ tiles and $158,811$ labels in the validation set.

\minisection{\xview clutter and imbalance.} Our version of \xview retains the drastic data imbalance and clutter with the top two categories containing over 100,000, 23 categories containing less than 1,000, and 7 categories containing less than 100 instances. With $20.6$ objects on average, the image tiles of \xview appear highly cluttered compared to the \coco dataset which has $7.7$ objects on average spread across larger images. \xview image tiles also reach a much higher \textit{maximum} clutter compared to \coco.  Whereas \coco images cap out at 15 object per image, \xview image tiles reach drastically higher levels of density: The top-10$\%$ of \xview image tiles contain an average of 101 objects and the most cluttered image tiles in xView contain over 500 objects.

\minisection{\coco~\cite{lin2014microsoft}}
We use the 2017 version of COCO which consists of $118,287$ images and $860,001$ objects in the training set, and $5,000$ images and $36,781$ objects in the test set. On average, the COCO dataset contains 7.7 object instances per image. This amount is less than would naturally occur since the strategies employed in the dataset curation explicitly avoided annotating more than 10-15 object instances within a single image  \cite{lin2014microsoft}. During training and testing, we resize the images with a minimum short side $704$ and a maximum long side $1333$ while keeping the aspect ratio.

\minisection{Performance measurement.} We measure the performance on the \xview and \coco datasets by calculating the Average Precision (AP) at an IoU threshold of 0.5.  This moderate threshold level is preferred over the stricter $0.75$ or $0.5:0.95$ thresholds because it is less sensitive to miniscule box shifts at the scale of the smallest object sizes in both datasets. Both datasets are evaluated in low data regimes in the early splits, e.g. ~100 labels per category, and a less stringent metric allows for better differentiation in challenging conditions.

\minisection{Object detector architecture and training details.} All model-based methods (D-MAL, entropy, and model-based random) for all approaches 
use a Faster-RCNN detector with a ResNet-50 backbone with Feature Pyramid Network (FPN). With the first two layers frozen, each FRCNN model is trained on a GeForce RTX 2080 Ti (11GB).  All experiments are trained $5$ times with unique random seeds and the mean reported.  The \coco experiments were trained for $30$ epochs with a batch size of $2$, learning rate $0.01$, using an SGD optimizer. 
The learning rate is reduced by 0.1 at 22 and 28 epochs. The \xview experiments were trained for 30 epochs with a batch size of $4$, learning rate $0.01$, using an SGD optimizer. 
The learning rate is reduced by 0.1 at 25 and 28 epochs.

In our experiments, we use four active learning methods among the three approaches.  These methods vary in their scoring functions and selection strategies.  We briefly describe each active learning method below and provide more approach-specific details in the next section, \ref{sec:approaches}. 
\vspace{-5pt}
\begin{enumerate}[leftmargin=*]
    \item \textbf{Minimax Active Learning for Detection (D-MAL).} We use our novel extension of MAL for object detection which was described in Section \ref{sec:mal4det}.
    \vspace{-5pt}
    \item \textbf{Maximum Entropy (MaxEnt)}  \cite{settles2014active} prioritizes objects with the highest entropy (in the Object-level approach) or images with the highest aggregated entropy score (in the Image-level approach) to be labeled.
    \vspace{-5pt}
    \item \textbf{Model-based Random (ModelRand)} randomly selects from the set of all objects predicted by the detection model. Note that no score is computed or employed in this method. Despite its simplicity, model-based random selection is surprisingly competitve in many active learning benchmarks. We use this baseline for only the Object-level and \method approaches.
    \vspace{-5pt}
    \item \textbf{Model-free Random (Random)} \textit{purely random} selection strategy used for the \image approach. Without any model input, images are selected at random for exhaustive labeling.
\end{enumerate}

\subsection{Active learning approaches setup}\label{sec:approaches}

In this section, we describe the setup of the two baseline approaches Image-level and Object-level, and our proposed Region-level approach, \method.  For each approach, we describe the oracle querying process, including any relevant distinctions between active learning selection methods to which the approaches are applied.

\minisection{Image-level.} The Image-level approach selects one image at a time to be exhaustively labeled by the oracle until the budget in terms of individual labels is reached, marking the end of a single \textit{split}. Each model-based selection method computes an Image-level informativeness score based on the object predictions made by the respective model. No score is computed for the Random baseline method. While the informativeness scores at the object level may differ across methods, the Image-level aggregation method is the same: we take the mean of all object informativeness scores to get a single Image-level score. The baseline methods evaluated in the Image-level approach are: Maximum Entropy, and Random. 

\minisection{Object-level.} The Object-level approach selects one predicted object instance at a time to be labeled by the oracle.  A label is only returned if a ground-truth instance of any category exists in its location that has any Intersection-over-Union (IoU) score of $0.25$, or greater, with the predicted query box. If there are multiple ground truth boxes intersecting the query, then the one with the highest IoU with the ground truth is selected regardless of the predicted category of the query.  
For each Object-level query to the oracle, one of two interactions occurs: either a ground-truth bounding box label is returned, or the oracle determines that no ground-truth bounding box label exists in the location of the predicted query box and no label is returned. Without accounting for the latter type of interaction, a vastly innefficient random query predictor would falsely exhibit comparable performance to a model that randomly selected among accurate predictions, hiding the fact that it could require orders of magnitude more interactions. 
The baseline methods evaluated in the Object-level approach are: Maximum Entropy, and Model-random.  Maximum Entropy is essentially the same as in Image-level, but without the Image-level aggregation. Each predicted object in the unlabeled pool is ranked and sent as queries to the oracle in order of descending predicted informativeness. Model-random selects randomly shuffled predictions made on the unlabeled pool to send as queries. 

\minisection{Region-level.} Our proposed Region-level approach, \method, selects one image region at a time to be labeled by the oracle. 
A single \textit{region} can encompass any single rectangular area within an image including, but not limited to, a single object or the entire image. 
The oracle returns all ground truth labels which are at least 25$\%$ covered by the region query box, regardless of the predicted category of the query. The budget is measured in terms of ground truth objects labeled by the oracle as well as region queries that yield zero groundtruth objects. The methods evaluated in the Region-level approach are: D-MAL, Maximum Entropy, and Model-random. The region scores are computed according to Section~\ref{sec:method}.  
Each region found in the unlabeled pool is ranked according the respective selection method's ranking algorithm and sent to the oracle for labeling in descending order. 

\begin{figure}[t]
\includegraphics[width=.46\textwidth]{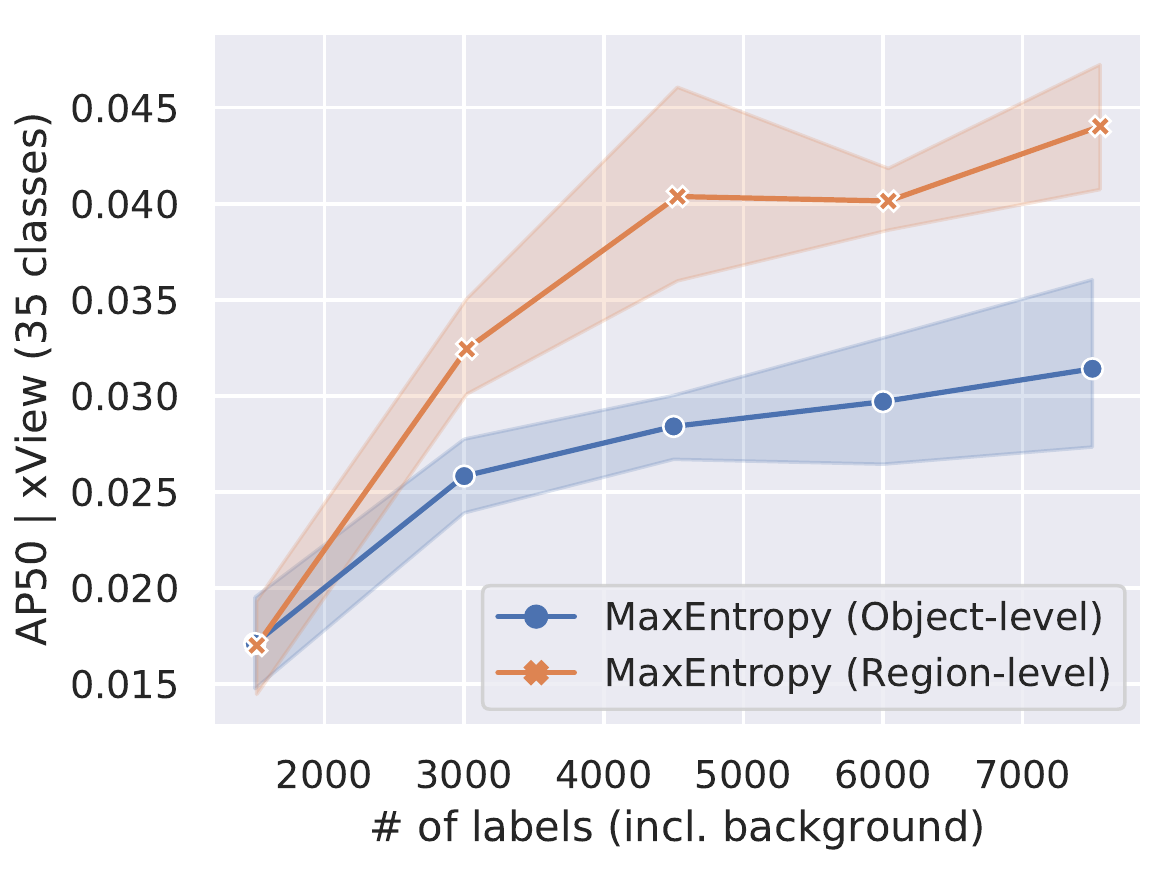}
    \vspace{-5pt}
    \caption{xView AP50 results for our splits in terms of the total labels applied by the oracle including background instances.  
    Our \ours approach significantly boosts the performance of entropy-based methods in cluttered scenes over an \object approach. Performance is low for both datasets since we're evaluating with relatively few instances per class and no data augmentation.
    }
    \label{fig:xview_ap50}
    \vspace{-2pt}
\end{figure}

\begin{figure}[t]
\includegraphics[width=.46\textwidth]{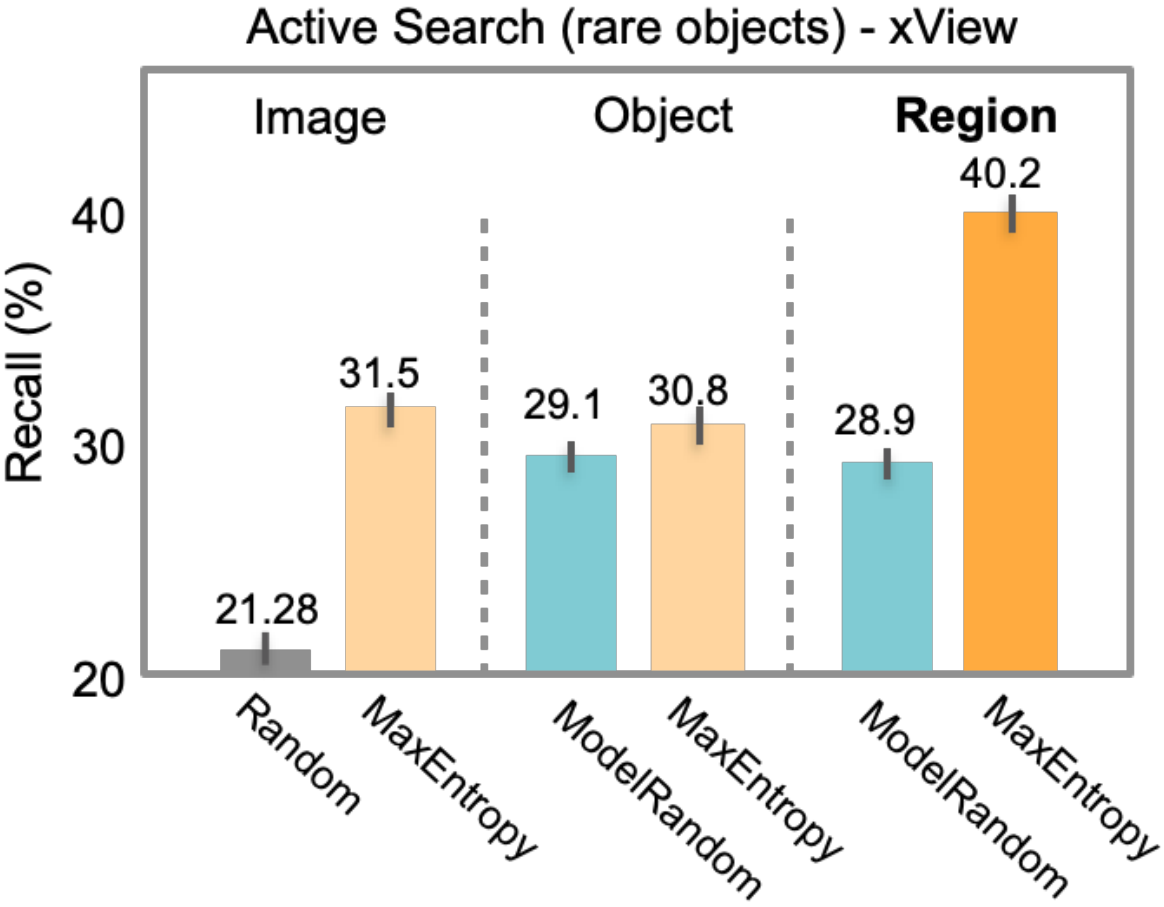}
\vspace{-5pt}
    \caption{Active search for rare objects on the xView dataset.  This plot shows the results after four active splits with a budget of 1500.  ReAL with MaxEntropy method outperforms Image-level and Object-level approaches by a significant margin.
    }
    \label{fig:xview_activesearch}
    \vspace{-10pt}
\end{figure}

\begin{table*}[ht]
\begin{center}
\footnotesize

\begin{tabular}{lccccc}
\toprule
\textbf{Approach (Method)} & \textbf{Split 0} & \textbf{Split 1} & \textbf{Split 2} & \textbf{Split 3} & \textbf{Split 4} \\
{\tiny \textit{Init. budget: 1500, Budget: 1500}} & {\tiny AP50 $\%$ (STD)} & {\tiny AP50 $\%$ (STD)} & {\tiny AP50 $\%$ (STD)} & {\tiny AP50 $\%$ (STD)} & {\tiny AP50 $\%$ (STD)} \\
\midrule
Image-level (Random) & 1.7 (0.3) & 2.0 (0.3) & 2.4 (0.2) & 2.0 (0.3) & 2.4 (0.2) \\
Image-level (MaxEnt) & 1.7 (0.3) & 2.0 (0.3) & 2.2 (0.2) & 3.1 (0.2) & 3.7 (0.2) \\
\midrule
Object-level (ModelRand) & 1.7 (0.3) & 2.9 (0.2) & 3.4 (0.2) & 3.9 (0.4) & 3.7 (0.3) \\
Object-level (MaxEnt) & 1.7 (0.3) & 2.6 (0.2) & 2.8 (0.2) & 3.0 (0.4) & 3.1 (0.5) \\
\midrule
ReAL (ModelRand) & 1.7 (0.3) & 3.0 (0.2) & 3.5 (0.2) & 3.7 (0.3) & 3.7 (0.3) \\
ReAL (MaxEnt) & 1.7 (0.3) & 3.2 (0.3) & 4.0 (0.6) & 4.0 (0.2) & 4.4 (0.4) \\
ReAL (D-MAL) & \textbf{2.8} (0.1) & \textbf{3.7} (0.3) & \textbf{4.2} (0.4) & \textbf{4.1} (0.2) & \textbf{4.6} (0.2) \\
\bottomrule
\end{tabular}
\caption{Active learning performance results across approaches and methods on the xView dataset. We use a budget of 1500 labels (initial and otherwise) and report the mean across five seeds with standard deviation in parentheses. \ours outperforms across all splits for both the Maximum Entropy (MaxEnt) and D-MAL methods.} \label{tab:xviewAP50}
\end{center}
\vspace{-10pt}
\end{table*}

\begin{table*}[ht]
\begin{center}
\footnotesize
\caption{Performance on the top-10 (most common), middle-10, and bottom-10 (rarest) category groupings in terms of total instances in the xView dataset. \ours outperforms the baselines across all splits in the middle-10 and rarest categories in terms of accuracy (AP50) and object search (Labeled $\%$).}
\label{tab:top-to-bottom-table}
\begin{tabular}{llllll}
\toprule
\textbf{Approach (Method)}        & \textbf{Split 0}      & \textbf{Split 1}      & \textbf{Split 2}      & \textbf{Split 3}      & \textbf{Split 4}      \\
\textit{Top-10 categories} & {\tiny AP50 $\%$ / Labeled $\%$} & {\tiny AP50 $\%$ / Labeled $\%$} & {\tiny AP50 $\%$ / Labeled $\%$} & {\tiny AP50 $\%$ / Labeled $\%$} & {\tiny AP50 $\%$ / Labeled $\%$} \\
Image-level (Random)     & 0.5 / 1.76 & 1.5 / 2.15 & 2.5 / 2.53 & \textbf{4.1} / 2.85 & 4.9 / 3.23 \\
Image-level (MaxEnt)     & 0.6 / 1.76 & 1.1 / 2.14 & 2.0 / 3.05  & 3.9 / 4.23 & \textbf{5.1} / 5.24 \\
Object-level (ModelRand) & 0.5 / 1.76 & 1.5 / 2.41 & 2.1 / 2.91 & 2.3 / 3.39 & 2.4 / 3.79 \\
Object-level (MaxEnt)    & 0.5 / 1.76 & 1.5 / 2.59 & 1.9 / 2.91 & 1.9 / 3.17 & 2.3 / 3.4  \\
ReAL (ModelRand)         & 0.5 / 1.76 & \textbf{1.8} / 2.73 & 2.3 / 3.43 & 3.0 / 4.32  & 3.2 / 4.93 \\
ReAL (MaxEnt)            & 0.5 / 1.76 & 1.7 / \textbf{4.01} & \textbf{2.7} / \textbf{5.75} & 3.2 / \textbf{6.88} & 4.2 / \textbf{7.79} \\
ReAL (D-MAL)              & 1.1 / 1.76 & 1.3 / 3.36 & 1.5 / 4.71 & 1.7 / 5.74 & 1.9 / 6.57 \\
\midrule
\textit{Middle-10 categories} &  &  &  & & \\
Image-level (Random)     & 4.3 / 7.53 & 4.8 / 7.92  & 5.2 / 8.2   & 5.7 / 8.62  & 6.2 / 8.95  \\
Image-level (MaxEnt)     & 4.2 / 7.53 & 4.5 / 8.19  & 5.0 / 9.18   & 6.1 / 10.57 & 7.1 / 12.5  \\
Object-level (ModelRand) & 4.1 / 7.53 & 6.7 / 10.69 & 7.6 / 11.95 & 8.0 / 13.14  & 8.1 / 14.2  \\
Object-level (MaxEnt)    & 4.3 / 7.53 & 5.7 / 9.76  & 6.4 / 10.9  & 6.2 / 11.65 & 6.6 / 12.26 \\
ReAL (ModelRand)         & 4.2 / 7.53 & 6.9 / 10.36 & 7.6 / 11.5  & 8.2 / 12.67 & 8.4 / 13.63 \\
ReAL (MaxEnt)            & 4.2 / 7.53 & 6.9 / 12.46 & 8.0 / 14.87  & 8.3 / 16.79 & 8.6 / 17.88 \\
ReAL (D-MAL)              & 6.6 / 7.53 & \textbf{7.9} / \textbf{13.11} & \textbf{8.7} / \textbf{16.3}  & \textbf{8.8} / \textbf{18.42} & \textbf{9.1} / \textbf{19.96} \\
\midrule
\textit{Bottom-10 categories} &  &  &  & & \\
Image-level (Random)     & 0.1 / 20.2  & 0.1 / 20.45 & 0.2 / 20.66 & 0.3 / 20.94 & 0.2 / 21.28 \\
Image-level (MaxEnt)     & 0.1 / 20.2  & 0.2 / 21.78 & 0.2 / 24.74 & 0.4 / 28.03 & 0.5 / 31.49 \\
Object-level (ModelRand) & 0.1 / 20.2 & 0.5 / 26.77 & 0.8 / 27.62 & 1.1 / 28.41 & 0.9 / 29.09 \\
Object-level (MaxEnt)    & 0.1 / 20.2 & 0.5 / 27.46 & 0.6 / 28.98 & 0.7 / 29.8  & 0.9 / 30.75 \\
ReAL (ModelRand)         & 0.1 / 20.2  & 0.5 / 23.03 & 0.4 / 25.16 & 0.5 / 27.2  & 0.6 / 28.9  \\
ReAL (MaxEnt)            & 0.1 / 20.2  & 0.5 / 29.43 & 0.8 / 33.48 & 1.0 / 37.34  & 0.9 / \textbf{40.18} \\
ReAL (D-MAL)              & 1.0 / 20.2   & \textbf{1.6} / \textbf{31.49} & \textbf{1.9} / \textbf{36.04} & \textbf{1.5} / \textbf{38.4}  & \textbf{2.2} / 40.16 \\
\bottomrule
\end{tabular}
\end{center}
\vspace{-15pt}
\end{table*}

\section{Results}

We discuss our results for all approaches and methods.

\minisection{xView.} We evaluate performance on the xView dataset \cite{xview}. The \method approach makes significant improvements in terms of both Average Precision (AP50) and Rare Object Search (ROS) over Image-level and Object-level approaches.  Figure \ref{fig:xview_ap50} demonstrates that the \ours approach can improve the performance of a model-based uncertainty method by reaching 
a $42\%$ relative improvement over the \object baseline (3.1\% to 4.4\%). \method achieves similar performance within the first split compared to what \object achieves by the final split i.e. \method can match the performance of \object with $60\%$ less labeling effort by the oracle. We see a striking difference in performance of the active learning approaches between \xview and \coco.  While \method performs similar to the baseline on \coco (details below), it outperforms on \xview, where heavy object clutter and class imbalance are prevalent.

Table~\ref{tab:xviewAP50} shows the performance of all approaches and methods on the \xview benchmark. The \ours approach outperforms across all splits for both the Maximum Entropy (MaxEnt) and D-MAL methods.  In the final split, \ours (D-MAL) improves over the strongest baseline, \object (ModelRand), by a 24$\%$ relative increase in AP50 (3.7\% to 4.6\%).  Furthermore, \ours (D-MAL) reduces the labeling effort by up to 50$\%$ considering it takes \object (ModelRand) until Split 3 to reach the accuracy achieved by \ours (D-MAL) in Split 1.

Table~\ref{tab:top-to-bottom-table} breaks down the results into the top-10 most common, bottom-10 rarest, and middle-10 categroies in the \xview dataset. The \ours approach outperforms the baseline approaches in the middle-10 and rarest categories in terms of ave. precision. \ours (D-MAL) shows a relative improvement over the strongest baseline of 12.3$\%$ in the middle-10 categories and 133$\%$ in the rarest categories. In terms of rare object search, \ours outperforms across all category breakdowns for both the Maximum Entropy (MaxEnt) and D-MAL methods. \ours (MaxEnt) finds 8.7$\%$ more rare objects (a 28$\%$ relative increase) over the strongest baseline (\image MaxEnt) in the last split. In the middle-10 categories, \ours (D-MAL) finds 5.8$\%$ more objects (a 41$\%$ relative increase) over the strongest baseline, \object (ModelRand), in the last split.  Surprisingly, even in the most common categories, \ours finds a higher percentage of the total instances in the dataset.  This can be explained by \image spending its labeling budget on the top-2 most common categories, \textit{buildings} and \textit{cars} which outnumber the other most common categories by an order-of-magnitude i.e. it cannot avoid oversampling these two categories. \ours avoids this pitfall and more than makes up the difference in the remaining 8 most common categories. 
\vspace{-5pt}
\minisection{COCO 2017.} Performance on the \coco benchmark is indistinguishable between \object and \ours. After 4 splits with a budget of 2500 both \ours and \object reach an Average Precision at 50\% IoU (AP50) of 3.3\%. This is somewhat expected since COCO is significantly less cluttered and more balanced across categories than the less-curated xView. This was our motivation for evaluating on the xView dataset with minimal pre-selection and highly cluttered images. 

\section{Conclusion}  \label{sec:discuss}

Existing approaches to active learning for detection adopt either an \image approach, whereby selected images are exhaustively labeled, or an \object approach, where single object candidates are sent to an oracle.  Both of these can be suboptimal in real-world situations.  We introduced a new Region-level approach that subsumes the Image- and Object-level approaches into a generalized approach that allows region-level selection and promotes spatial-diversity by avoiding nearby redundant queries from the same image.  In addition, we developed a novel detector variant of a recently introduced minimax active learning method that balances diversity and uncertainty measures. We show that our approaches significantly decrease labeling effort on natural, real-world data with inherent class-imbalance and cluttered scenes.


{\small
\bibliographystyle{ieee_fullname}
\bibliography{main}
}

\appendix

\setcounter{table}{2}
\twocolumn[
\begin{center}
\Large 
\textbf{Region-level Active Detector Learning}
\\
(Supplementary Materials)
\par
\end{center}
\vspace{10pt}
]


\section{Breakdown of active learning methods}

\begin{table}[ht]

\caption{We evaluate \method against Image- and Object-level baseline approaches on four active learning methods of various scoring functions and selection strategies.  Two types of random methods are included, one that samples randomly over model object predictions and the other samples entire images at random.}
\begin{center}
\small
\begin{tabular}{| p{1.5cm} | p{2.2cm}| p{1.7cm} | p{0.9cm} |}
\hline
     & Scoring  & Selection & Model \\
Method & function & strategy  & based  \\
\hline\hline
MAL & Entropy, $E$; & Min. $L$, & yes \\
 & Labeledness, $L$ & Max. $E$ & \\
 \hline
Maximum & Entropy, $E$ & Max. $E$ & yes\\
Entropy &  &  & \\
\hline
Model- & None & Random & yes\\
random &  & selection & \\
\hline
Random & None & Random & no\\
 &  & selection & \\
\hline
\end{tabular}
\end{center}
\label{tab:methods}
\end{table}

\section{xView details}
\minisection{Image resolution.} \xview images are captured on MAXAR satellite platforms that image the Earth from an altitude of $617$km.  Despite featuring the highest resolution of any commercial platform, 0.3-0.5m, many object categories appear at low resolutions due to their relatively small physical size. In addition, MAXAR satellite images undergo a post-processing to adjust for geospatial registering and panchromatic sharpening, which adds more pixel noise and artifacts. 

\minisection{Image redundancy.} In addition to extreme class imbalance, \xview exhibits image redundancy rarely captured in common benchmarking datasets. Since the original \xview images are quite large, a significant portion of the small tiles will contain `dead' space in the form of images over uniform terrain such as open sea or desert which contain little, if any, objects of interest. 

This aspect of \xview captures the realistic quality that data is almost always over-collected. That is, much more data gets collected than is needed or that could be practically labeled. Active learning becomes substantially more valuable in this case, since (purely) random sampling will do nothing to avoid selecting such redundant images.

\minisection{Class imbalance.} See table \ref{tab:xview-counts} for a list of xView category names with corresponding instance and image counts.

\begin{table}[ht]
\begin{center}
\caption{The number of instances and images for each of the 35 categories in xView (descending order by \textit{Number of instances}).}
\label{tab:xview-counts}
\begin{tabular}{lll}
\toprule
\textbf{Category}       & \textbf{\# of} & \textbf{\# of} \\
\textbf{name}       & \textbf{instances} & \textbf{images} \\
\midrule
Other Building               & 233098                       & 8526                      \\
Car                          & 140647                       & 6650                      \\
Other Truck                  & 15556                        & 4038                      \\
Truck w/Trailer Bed          & 6459                         & 1413                      \\
Bus                          & 4618                         & 1684                      \\
Other Passenger Vehicle      & 2062                         & 426                       \\
Other Maritime Vessel        & 1721                         & 400                       \\
Other Railway Vehicle        & 1605                         & 137                       \\
Motor/Sail/Small Boat        & 1453                         & 196                       \\
Dump/Haul Truck              & 1241                         & 510                       \\
Passenger Car                & 1172                         & 101                       \\
Storage Tank                 & 1137                         & 318                       \\
Facility                     & 879                          & 584                       \\
Shed                         & 771                          & 435                       \\
Other Engineering Vehicle    & 758                          & 400                       \\
Damaged Building             & 728                          & 387                       \\
Excavator                    & 665                          & 368                       \\
Passenger/Cargo Plane        & 583                          & 332                       \\
Loader/Dozer/Tractor/Scraper & 518                          & 315                       \\
Hut/Tent                     & 471                          & 126                       \\
Pylon                        & 314                          & 239                       \\
Container Ship               & 284                          & 184                       \\
Small Plane                  & 217                          & 72                        \\
Aircraft Hangar              & 200                          & 108                       \\
Barge                        & 181                          & 100                       \\
Tower Crane                  & 144                          & 96                        \\
Crane Truck                  & 141                          & 100                       \\
Container Crane              & 132                          & 78                        \\
Helipad                      & 92                           & 47                        \\
Locomotive                   & 84                           & 58                        \\
Truck w/Liquid Tank          & 82                           & 54                        \\
Tower Structure              & 63                           & 52                        \\
Straddle Carrier             & 55                           & 28                        \\
Other Aircraft               & 47                           & 35                        \\
Helicopter                   & 39                           & 23   
\end{tabular}
\end{center}
\end{table}

\end{document}